\documentclass[sigconf]{acmart}
\AtBeginDocument{%
  }

\usepackage{booktabs}
\usepackage{multirow}
\usepackage[normalem]{ulem}
\useunder{\uline}{\ul}{}
\usepackage{float}  

\usepackage{subcaption}

\usepackage{amsmath, amssymb}
\usepackage[ruled,linesnumbered]{algorithm2e}

\usepackage{wrapfig}

\begin{document}

\title{WebRenderBench: Enhancing Web Interface Generation through Layout-Style Consistency and Reinforcement Learning}

\author{Peichao Lai}
\authornote{Both authors contributed equally to this research.}
\affiliation{%
  \institution{Peking University}
  \city{Beijing}
  \country{China}
}

\author{Jinhui Zhuang}
\authornotemark[1]
\affiliation{%
  \institution{Xiamen Huaxia University}
  \city{Xiamen}
  \state{Fujian}
  \country{China}}

\author{Kexuan Zhang}
\affiliation{%
  \institution{Fuzhou University}
  \city{Fuzhou}
  \state{Fujian}
  \country{China}
}

\author{Ningchang Xiong}
\affiliation{%
 \institution{City University of Hong Kong}
 \city{Hong Kong SAR}
 \country{China}}

\author{Shengjie Wang}
\affiliation{%
  \institution{Fuzhou University}
  \city{Fuzhou}
  \state{Fujian}
  \country{China}}

\author{Yanwei Xu}
\affiliation{%
  \institution{Peking University}
  \city{Beijing}
  \country{China}}

\author{Chong Chen}
\affiliation{%
  \institution{Huawei Cloud BU}
  \city{Beijing}
  \country{China}}

\author{Yilei Wang}
\affiliation{%
  \institution{Fuzhou University}
  \city{Fuzhou}
  \state{Fujian}
  \country{China}}
\email{yilei@fzu.edu.cn}

\author{Bin Cui}
\affiliation{%
  \institution{Peking University}
  \city{Beijing}
  \country{China}}
\email{bin.cui@pku.edu.cn}

\settopmatter{printacmref=false} 

\renewcommand{\shortauthors}{Peichao Lai et al.}

\begin{abstract}
  Automating the conversion of UI images into web code is a critical task for front-end development and rapid prototyping. Advances in multimodal large language models (MLLMs) have made WebUI-to-Code increasingly feasible, yet existing benchmarks remain limited in data diversity and evaluation reliability. To address these issues, we present WebRenderBench\footnote{https://huggingface.co/datasets/aleversn/WebRenderBench}, a large-scale benchmark of 45.1k webpages collected from real-world portal sites, offering greater diversity, complexity, and realism than prior benchmarks. We further propose a novel evaluation metric that measures layout and style consistency from the final rendered pages. Unlike vision-based methods that rely on costly LLM reasoning or structure-based comparisons vulnerable to noise and asymmetry, our approach enables more efficient, objective, and reliable UI quality assessment. Finally, we introduce the Automated Layout and Style Inspection Agent (ALISA), which integrates this metric into reinforcement learning as a reward signal to enhance training on crawled asymmetric webpages. Experiments show that ALISA significantly boosts generation performance, achieving state-of-the-art results across multiple metrics.
\end{abstract}



\begin{teaserfigure}
  \includegraphics[width=\textwidth]{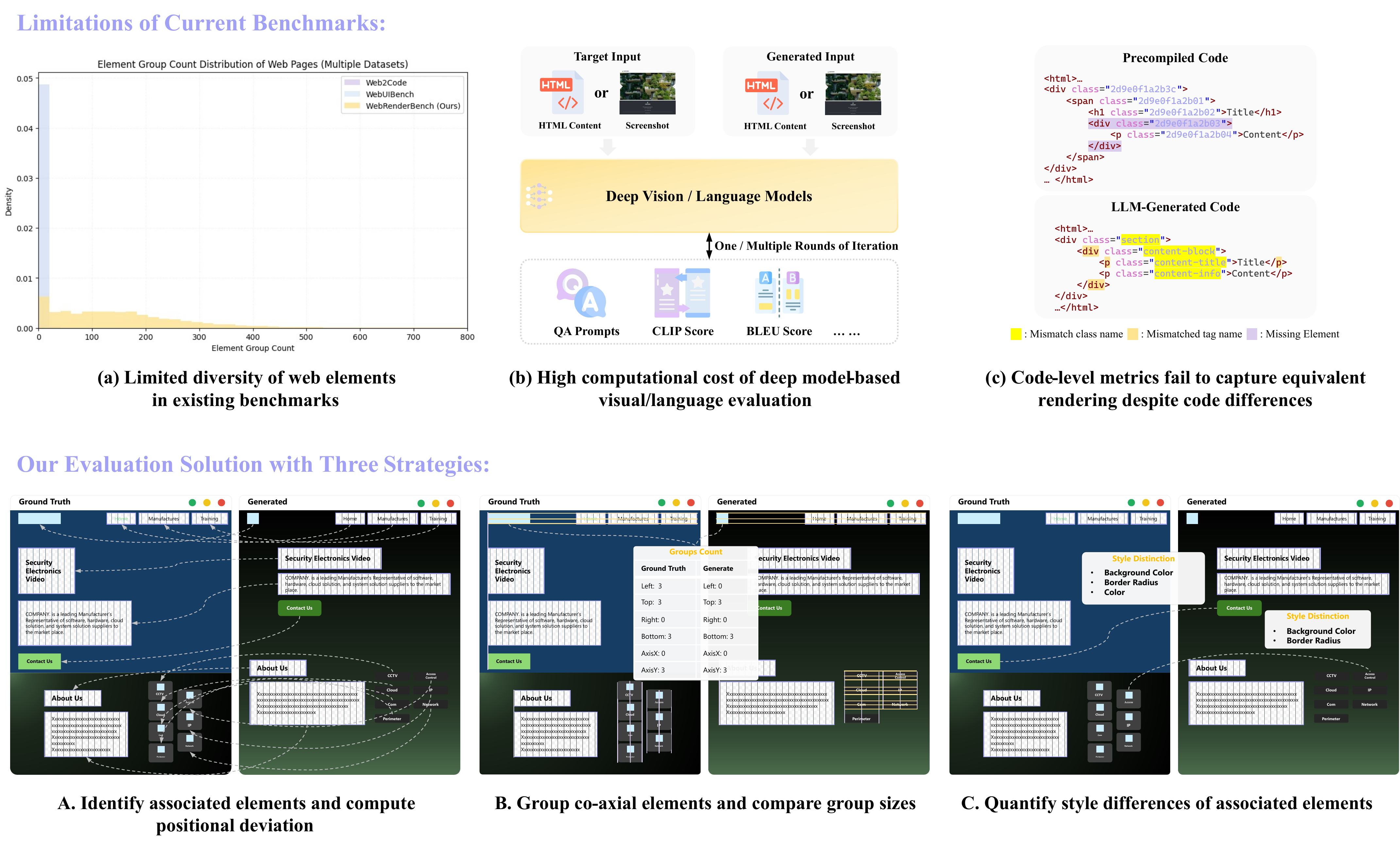}
  \caption{Limitations of current WebUI-to-Code benchmarks and illustrative examples of our proposed solutions.}
  \Description{Enjoying the baseball game from the third-base
  seats. Ichiro Suzuki preparing to bat.}
  \label{f1}
\end{teaserfigure}


\maketitle

\section{Introduction}
Automating the conversion of UI images into web code holds great significance for both front-end development and automated prototyping \cite{DBLP:journals/corr/abs-2405-04975, xiao2025interaction2codebenchmarkingmllmbasedinteractive}. The advent of multimodal large language models (MLLMs) \cite{Qwen-VL, NEURIPS2023_6dcf277e, wang2025internvl3_5} have brought notable advances in code generation, making WebUI-to-Code increasingly feasible. Nevertheless, current MLLMs still face limitations in generation quality, and the establishment of reliable benchmarks and evaluation metrics remains essential for driving progress in this field.

The goal of the WebUI-to-Code task is to generate accurate HTML code from user-provided UI images, ensuring that the resulting layout and style closely align with the input design. To assess the WebUI-to-Code capabilities of MLLMs, several benchmarks \cite{DBLP:conf/nips/YunLTBWJDWTL0NB24, DBLP:conf/naacl/SiZLYLY25, DBLP:conf/acl/LinZ0WM0025, DBLP:conf/eics/Beltramelli18} have been introduced that offer paired images, reference HTML code, and corresponding evaluation metrics. However, as shown in Figure~\ref{f1}, existing benchmarks still exhibit several notable limitations:

\textbf{(1) Limitations in data quality.} Existing benchmark evaluation sets typically contain webpages with relatively simple structures and limited scale. Moreover, datasets such as Web2Code \cite{DBLP:conf/nips/YunLTBWJDWTL0NB24} are primarily composed of webpages synthesized by LLMs, which restricts data diversity. As shown in Figure~\ref{f1}(a), we compare the diversity of webpage elements across existing datasets and our dataset using the number of grouped elements (see Algorithm~\ref{a2}). The results show that existing datasets generally exhibit lower element diversity, which limits their ability to reflect model performance on complex, real-world UI images.

\textbf{(2) Limitations in evaluation capability.} Current approaches for automatically evaluating generated webpages can be broadly categorized into two types. The first is vision-based evaluation, as adopted in datasets such as Web2Code \cite{DBLP:conf/nips/YunLTBWJDWTL0NB24} and Design2Code \cite{DBLP:conf/naacl/SiZLYLY25}. In this approach, generated webpages are rendered into screenshots and compared with ground-truth UI images using large-scale models such as GPT-4 Vision API. Multi-turn question answering is then applied to assess element consistency. However, this method entails substantial computational overhead, suffers from low efficiency due to repeated QA interactions, and cannot directly measure layout and style consistency. Instead, it relies heavily on the visual reasoning abilities of large models, which may overlook subtle element differences. The second type is structure-based evaluation. For instance, WebUIBench \cite{DBLP:conf/acl/LinZ0WM0025} assesses DOM tree similarity by matching corresponding elements in the generated and reference code based on class names and tag names. As illustrated in Figure~\ref{f1}(c), a single UI design can often be implemented using multiple valid code structures. When mismatches occur in nesting relations or tag usage between generated and reference code, this approach fails to accurately reflect generation quality. Moreover, when training webpages are obtained from crawled external sources, the retrieved code may include compiled class names and irrelevant tags, which we define as code asymmetry. Direct comparison with such noisy code further undermines the reliability of performance evaluation.

In this paper, we propose WebRenderBench to address the aforementioned challenges by improving both data quality and evaluation methodology. \textit{To overcome the limitations of existing datasets}, we construct a large-scale dataset sourced from real-world portal websites, comprising 45.1k webpages with greater diversity, complexity, and realism, thereby covering a broader spectrum of web designs. \textit{To alleviate the issue of evaluation capability}, we introduce a novel metric grounded in layout and style consistency. This metric leverages WebDriver-rendered outputs and spatial information from the final rendered pages. As illustrated in Figure~\ref{f1}(A–C), it compares generated webpages with their ground-truth counterparts by (i) matching associated elements and quantifying spatial discrepancies, (ii) evaluating grouping differences based on the number of aligned elements along the same axes, and (iii) measuring style differences through quantifiable attributes of corresponding elements. This approach effectively mitigates the noise and asymmetry commonly present in synthesized or crawled webpages, which enables more efficient, objective, and reliable evaluation of final UI quality. Furthermore, to improve model generation performance on externally obtained, asymmetric webpage data, we propose the \textbf{A}utomated \textbf{L}ayout and \textbf{S}tyle \textbf{I}nspection \textbf{A}gent (\textbf{ALISA}) for LLM-generated WebUIs. This framework allows our evaluation metric to serve not only as an offline evaluation tool but also as a reward signal within reinforcement learning, thereby enhancing the ability of vision language models (VLMs) to generate high-quality WebUI-to-Code outputs. Experimental results demonstrate that integrating ALISA significantly enhances GRPO training on crawled asymmetric webpages, leading to state-of-the-art performance across multiple evaluation metrics. We summarize our contributions as follows:

1. \uline{\textit{New Benchmark.}} We construct a large-scale, diverse, and realistic dataset of UI images paired with corresponding code, enabling more comprehensive evaluation of UI-to-Code generation.

2. \uline{\textit{New Evaluation Metrics.}} We propose a code-level metric based on layout and style consistency from the final rendered pages, overcoming the noise in code and the inefficiency of LLM-based evaluation.

3. \uline{\textit{State-of-the-art Performance.}} Our evaluation metric can be directly used as a reward signal in reinforcement learning, enhancing the generation capabilities of VLMs and achieving state-of-the-art performance on multiple evaluation metrics.

\begin{figure*}[ht]
  \includegraphics[width=\textwidth]{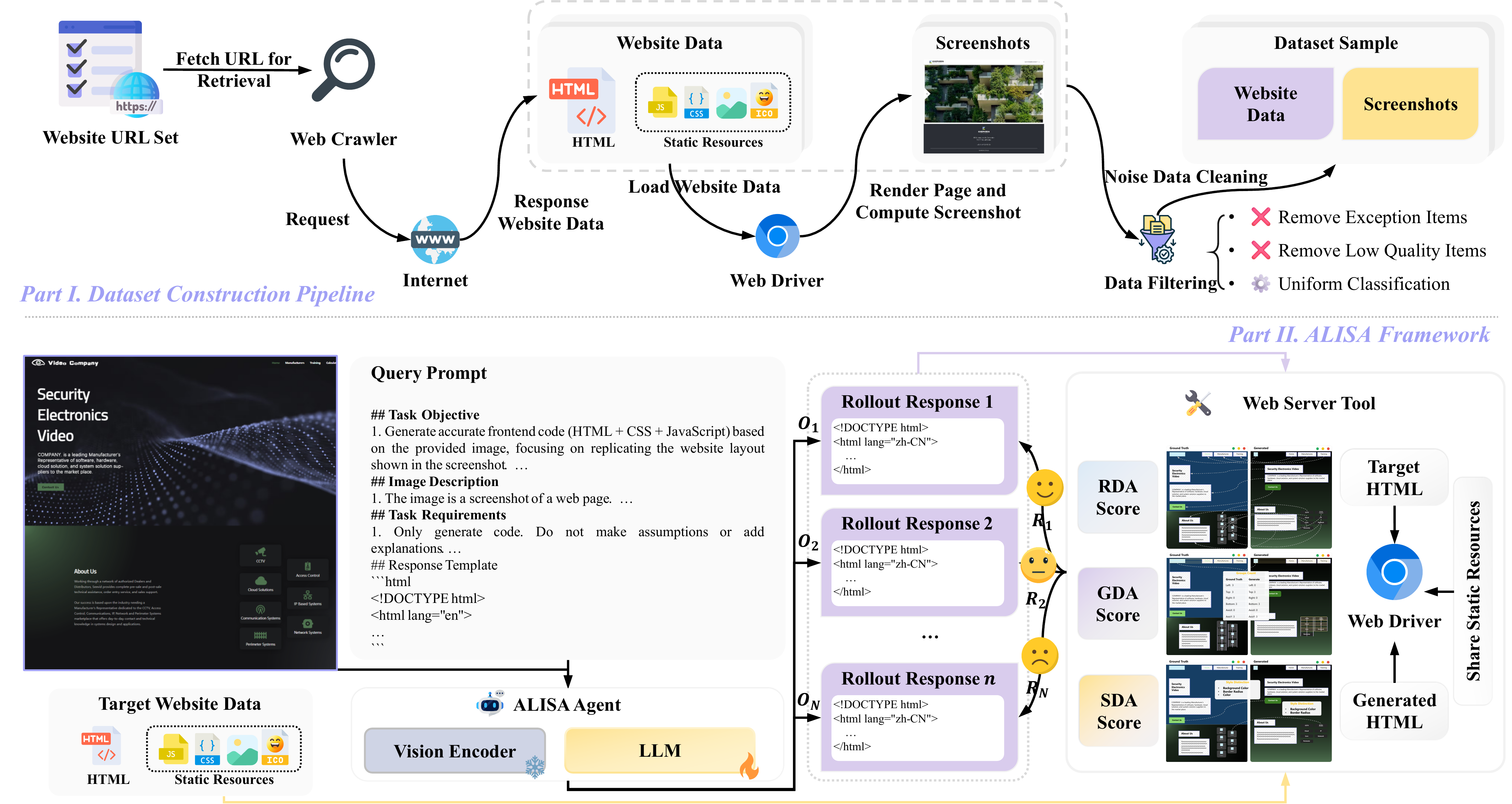}
  \caption{Dataset construction pipeline and the architecture of the ALISA framework.}
  \label{f2}
\end{figure*}

\section{Related Work}

\subsection{WebUI-to-Code Benchmarks and Evaluation Metrics}

Early studies on WebUI-to-Code employed deep learning models \cite{DBLP:conf/eics/Beltramelli18, DBLP:journals/tse/LiuSZNLZ22, DBLP:journals/tse/MoranBCBP20, DBLP:journals/smr/XuBSLJZ21}, which translated UI screenshots into a simplified domain-specific language using CNN–RNN architectures. These models were trained on synthetic, programmatically generated datasets and evaluated with text-matching metrics such as Jaccard and BLEU scores \cite{su2025crossingrewardbridgeexpanding, DBLP:journals/corr/abs-2009-10297, soselia2023learninguitocodereversegenerator}. However, they struggled to generalize due to the substantial gap between synthetic and real webpages. With the advent of MLLMs \cite{Qwen-VL, DBLP:journals/corr/abs-2312-14238}, which demonstrate strong zero- and few-shot generation capabilities, the focus has shifted toward benchmarks built on real-world data. Design2Code \cite{DBLP:conf/naacl/SiZLYLY25} introduced the first large-scale test set of 484 real webpages and replaced rigid code-matching metrics with vision-based measures such as CLIP Score \cite{DBLP:conf/emnlp/HesselHFBC21} and component-level fidelity. WebSight \cite{DBLP:journals/corr/abs-2403-09029} introduced a synthetic dataset of 2 million pairs of HTML code and screenshots. Similarly, Sketch2Code \cite{DBLP:conf/naacl/LiZY25} introduced a dataset of 731 examples, designed to evaluate both direct generation from low-fidelity inputs and multi-turn refinement through simulated user feedback and agent-initiated queries. WebUIBench \cite{DBLP:conf/acl/LinZ0WM0025} further broadened the scope by framing tasks from a software engineering perspective. Overall, modern benchmarks increasingly emphasize vision-level metrics, which better capture user experience and tolerate diverse yet functionally equivalent code implementations, thereby overcoming the limitations of purely text-based evaluation.

\subsection{Reinforcement Learning for Optimizing Code Generation}

While MLLMs trained with supervised fine-tuning (SFT) can generate structurally plausible code, they often struggle with layout and styling. Recent work tackles this alignment challenge with reinforcement learning, which goes beyond standard loss functions by optimizing objectives through preference-based strategies such as PPO \cite{DBLP:journals/corr/SchulmanWDRK17}, DPO \cite{DBLP:conf/nips/RafailovSMMEF23}, and GRPO \cite{DBLP:journals/corr/abs-2402-03300}. In code-related RL tasks, the model is treated as an agent \cite{DBLP:journals/corr/abs-2505-00753}, where each action corresponds to generating \cite{DBLP:journals/corr/abs-2506-03136, DBLP:journals/tmlr/LiuZXF00Y23, DBLP:journals/corr/abs-2502-01715} or refining code snippets \cite{11029811, DBLP:journals/corr/abs-2505-03335}. The environment (e.g., a compiler or runtime checker) provides a reward signal, which the model uses to update its parameters and maximize performance. The effectiveness of RL largely depends on reward design. In WebUI-to-Code tasks, three major reward schemes are commonly employed: (i) visual similarity between rendered outputs and target UIs, typically measured by CLIP or feature distance \cite{DBLP:conf/nips/GaoHXX23, DBLP:journals/corr/abs-2405-04975}; (ii) code quality and functionality, such as syntax correctness or unit test success \cite{DBLP:conf/naacl/LiZY25}; and (iii) human or AI feedback, aligning outputs with subjective preferences such as readability or aesthetics \cite{DBLP:conf/naacl/0001SLBB024}. These RL-based methods enable models to learn directly from feedback, moving beyond passive imitation toward generating code that is both visually faithful and functionally reliable.

\section{Benchmark Construction}

The overall pipeline for WebRenderBench data collection and the ALISA training framework are illustrated in Figure~\ref{f2}. $\S$~\ref{s31} outlines the core principles guiding our benchmark design. $\S$~\ref{s32} describes the dataset construction process and presents key statistics. Finally, $\S$~\ref{s33} introduces the evaluation methodology, defines the metrics used in our benchmark, and explains how ALISA is incorporated into the training framework.

\subsection{Benchmark Design}
\label{s31}

The motivation for our benchmark design is driven by three core questions:
(1) How can we effectively evaluate the WebUI-to-Code generation capabilities of LLMs in realistic and diverse application scenarios?
(2) How can we achieve efficient and accurate evaluation of generation quality at the code level?
(3) How can we design a reward mechanism that enables effective optimization when the ground-truth and generated code are asymmetric?
To address these challenges, we first design a data collection pipeline that acquires webpage data from real-world portal sites. After systematic cleaning and filtering, we construct a large-scale dataset with high diversity and a balanced distribution of webpage complexity. Next, we propose an evaluation method that employs a WebDriver with embedded JavaScript scripts to analyze layout and style consistency between generated and ground-truth pages based on their final rendered outputs. This approach not only ensures efficient and accurate evaluation but also resolves the asymmetry between generated code and ground-truth compiled code, enabling the evaluation scores to be directly used as rewards for optimizing the WebUI-to-Code capabilities of LLMs. We provide detailed discussions of these components in the following sections.

\begin{table}[ht]
\caption{Comparison of dataset statistics for WebUI-to-Code. In Web2Code, the test set is reported using image information and QA pairs, whereas WebRenderBench is categorized based on the Group Count values.}
\label{t1}
\resizebox{\columnwidth}{!}{%
\begin{tabular}{@{}lccccc@{}}
\toprule
\multirow{2}{*}{\textbf{Dataset}} & \multirow{2}{*}{\textbf{Web2Code} \cite{DBLP:conf/nips/YunLTBWJDWTL0NB24}} & \multirow{2}{*}{\textbf{WebUIBench} \cite{DBLP:conf/acl/LinZ0WM0025}} & \multicolumn{3}{c}{\textbf{WebRenderBench}}           \\
                                  &                                    &                                      & (0$\sim$200)    & (200$\sim$400)   & ($\geq$400)           \\ \midrule
\textbf{Source}                   & Synthetic                          & Real-World                           & Real-World      & Real-World       & Real-World       \\
\textbf{Train Size}                & 884.7k                               & -                                 & 11.4k           & 8.6k             & 2.6k             \\
\textbf{Eval Size}                & 1.2k                               & 4.5k                                 & 11.4k           & 8.5k             & 2.6k             \\
\textbf{Avg. Length}              & 1412.02±498.10                     & 1037.98±876.01                       & 7641.15±4643.40 & 11388.35±4968.60 & 11964.56±5439.50 \\
\textbf{Avg. Tag Count}           & 20.31±8.51                         & 8.96±5.40                            & 446.78±760.93   & 1239.96±1615.30  & 2505.50±2939.96  \\
\textbf{Avg. DOM Depth}           & 3.64±0.75                          & 2.55±0.68                            & 9.56±3.88       & 14.50±4.69       & 16.40±5.03       \\
\textbf{Avg. Group Count}         & 9.23±3.76                          & 3.88±3.08                            & 99.73±58.02     & 279.05±52.80     & 649.07±660.10    \\ \bottomrule
\end{tabular}%
}
\end{table}

\begin{figure}[ht]
  \includegraphics[width=\columnwidth]{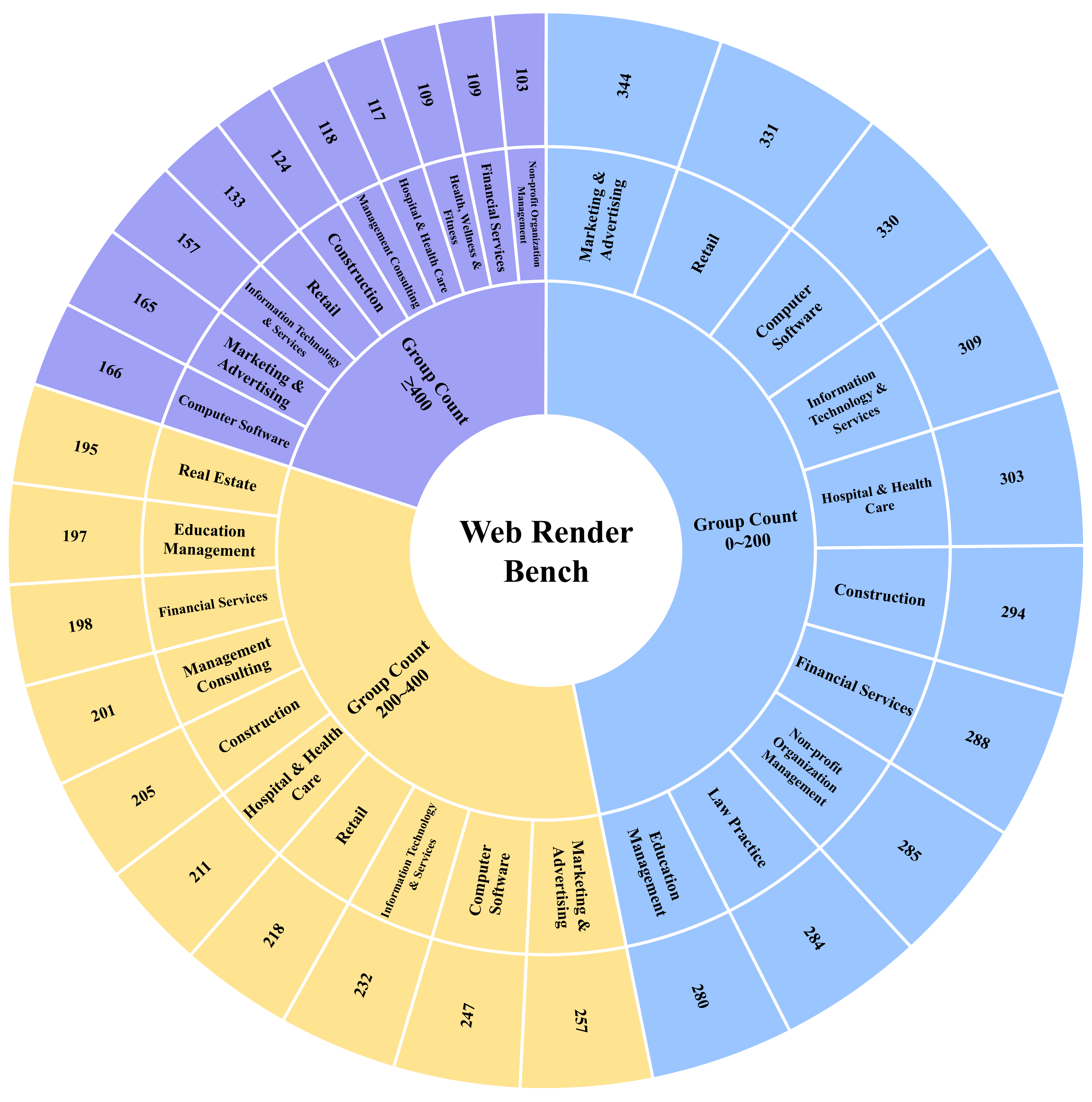}
  \caption{Sunburst chart showing the distribution of the top 10 industries by sample count across different Group Counts in our WebRenderBench test set.}
  \label{f3}
\end{figure}

\subsection{Dataset}
\label{s32}

Following the benchmark design, the WebRenderBench dataset is constructed through a systematic pipeline encompassing data collection, processing, and cleaning.

\subsubsection*{\textbf{Data Collection.}} We collected website URLs from an open-source corporate portal profile dataset\footnote{https://huggingface.co/datasets/SaleleadsOrg/linkedin-company-profile} and implemented a high-concurrency crawler to fetch the homepage HTML along with all associated static resources. We attempted to access about 350k portal websites and successfully retrieved 210k, which were initially stored in MHTML format.

\subsubsection*{\textbf{Data Processing.}} To support subsequent webpage parsing and evaluation, we post-processed the collected raw data. First, the compressed MHTML pages were converted into HTML files along with their corresponding static resource directories. During this process, we identified some resources that remained cross-origin and unavailable locally, which could hinder evaluation and training due to network dependencies. To address this, for each page, we traversed all cross-origin media resources, such as images and videos, recording their width, height, and aspect ratio. We also record the corresponding information for all local media and constructed a shared static resource library. Cross-origin resources were then replaced with local media of matching aspect ratios to mitigate potential performance issues. Next, using a browser driver, we opened each page and captured screenshots of the fully rendered content. Pages were initially loaded in a standard 1920×1080 viewport, and once the document fully loaded, the browser height was adjusted based on the page's scroll height to obtain full-page screenshots.

\subsubsection*{\textbf{Data Cleaning.}} Some locally obtained webpages contained missing resources or inaccessible CSS and JS files due to cross-origin restrictions, which could affect subsequent evaluation and training. The data also exhibited uneven distributions in element counts and industry domains, potentially introducing biases. To address these issues, we conducted systematic data cleaning. First, pages were filtered based on screenshot dimensions, removing excessively long or abnormally sized pages (height capped at 5,000 pixels) to prevent input overflow for vision-language models. Next, pages with rendering anomalies or missing style information were excluded. We observed that most style-deficient elements were positioned as static, aligned to the left edge, and lacked font styling. Following this pattern, we traversed all elements with ``left = 0'' and computed the proportion of elements with ``position: static'' and empty font styles to assign a style quality score to each page, discarding those with scores above 0.9. Additionally, we applied Qwen2.5-VL-7B-Instruct \cite{qwen2.5-VL} to perform QA on all screenshots to detect large blank areas or overlapping elements, further removing pages with missing elements or style errors. After these steps, the cleaned dataset comprised 110k pages. Finally, pages were categorized by industry and element complexity, measured via Group Count (see Algorithm~\ref{a2}), which more accurately reflects element diversity than simple element counts by mitigating the influence of repeated list-type elements.

\subsubsection*{\textbf{Data Statistics.}} We summarize the final dataset statistics in Table~\ref{t1}. After data cleaning, the dataset comprises 45.1k samples, evenly split into training and test sets. Compared to Web2Code and WebUIBench, our dataset contains a substantially larger number of test samples. Moreover, based on average label counts, DOM depth, and Group Count, it exhibits the highest information richness. The high standard deviation compared to other datasets indicates greater diversity, while the sample distribution in the sunburst chart (Figure~\ref{f3}) and in Figure~\ref{f1}(a) demonstrates that our dataset maintains a balanced complexity distribution.

\subsection{Protocol for Evaluation and Training}
\label{s33}

Existing benchmarks for WebUI-to-Code either evaluate code generation quality from the perspective of code structure or rely solely on visual comparisons of rendered webpages. Both approaches have notable limitations. Structure-based evaluation cannot effectively handle asymmetric code scenarios, while vision-based evaluation often provides only coarse-grained results. Although incorporating LLM-based QA or manual assessment into vision-based methods can improve accuracy, it also incurs substantial time costs, making them unsuitable for real-time reward computation during online training. To address these challenges, we leverage real-time DOM information from browser rendering and assess generation quality from both the rendered output and the code level. Building on the core objectives of WebUI-to-Code, namely layout consistency and style consistency, we propose three evaluation strategies:

\subsubsection*{\textbf{Relative Layout Difference of Associated Elements (RDA)}} For each visible element, high-quality code generation should ensure the existence of a corresponding element in the ground-truth target, with both size and spatial layout closely aligned. Following this principle, the first step is to establish associations between elements. Formally, let the set of visible elements from the generated code be $S = \left\{s_1, \ldots, s_n\right\}$, and the set of visible elements from the ground-truth target code be $T = \left\{t_1, \ldots, t_m\right\}$. To address inconsistencies in class or tag names, candidate associations are ranked based on inner-text similarity and position/size differences. For elements with inner text, we compute similarity using the Longest Common Subsequence (LCS) and directly match pairs with a score above 0.80. If multiple candidates satisfy this condition, the pair with the smallest position and size difference is selected. For elements without inner text, we instead rely on positional and size differences. We further introduce a threshold of ten based on common UI design padding bias\footnote{https://developer.apple.com/design/human-interface-guidelines/}: if the minimum size difference exceeds this threshold, the element is considered to have no valid association. Next, for each associated pair $(s, t)$, the RDA score is computed following Algorithm~\ref{a1}. We first divide each webpage into three horizontal quadrants (left, center, right) and three vertical quadrants (top, middle, bottom). An element is assigned to a horizontal quadrant depending on whether it lies entirely to the left of the center line, spans the center line, or lies entirely to the right; the same principle applies vertically. The spanning attribute may coexist with either of the other two. To identify major positional deviations, we strictly compare the quadrant assignments of $(s_i, t_j)$. If they differ, the pair receives a score of 0; otherwise, their relative deviation is computed using half of the window’s width and height as the reference threshold. To account for the varying importance of elements, each associated pair is weighted by its uniqueness. The final RDA score is then computed as the weighted sum of all associated pairs. Details of the weight calculation are presented in the GDA section.

\begin{algorithm}[h]
\caption{RDA Computation for Each Associated Pair}\label{a1}
\KwIn{Generated elements $S=\{s_1,...,s_n\}$, target elements $T=\{t_1,...,t_m\}$, race weight $w$, window width $h$, window height $v$}
\KwOut{RDA scores for associated elements}

\BlankLine
\textbf{Function: posSim(val1, val2, ref)} \\
\eIf{$|val1 - val2| / ref > 1$}{\Return 0}{\Return $1 - |val1 - val2| / ref$}

\BlankLine
\ForEach{associated pair $(s,t)$}{
    score $\gets 100 \cdot w$\;

    $(bx_s, by_s) \gets$ getQuadrant($s$)\;
    $(bx_t, by_t) \gets$ getQuadrant($t$)\;
    matchX $\gets$ compareBias($bx_s, bx_t$)\;
    matchY $\gets$ compareBias($by_s, by_t$)\;

    \If{not matchX or not matchY}{score $\gets 0$}

    score $\gets$ score $\cdot$ posSim($s.left, t.left, h / 2$) $\cdot$ posSim($s.top, t.top, v / 2$)\;

    assign(($s, t$), score)
}
\end{algorithm}

\subsubsection*{\textbf{Group-wise Difference in Element Counts (GDA)}} Beyond fine-grained evaluation of individual associated elements, we also consider the consistency of elements aligned along the same axes. In UI layout design, lists, grids, and tables typically align their items horizontally or vertically, and navigation bars or main content sections follow similar alignment patterns. Based on this observation, we propose GDA to measure the consistency in the number of grouped elements between the generated and target pages. Formally, for each element $e_i$, we define its reference axis set as $\mathcal{A}_i = \left\{ A_l, A_t, A_r, A_b, A_h, A_v \right\}$, representing its left, top, right, and bottom boundaries, as well as its horizontal and vertical center lines. The group $G_i$ associated with $e_i$ is then defined as:
\begin{align}
G_i = \left\{ e_j \in \text{Page} \mid \exists A \in \mathcal{A}_i, \mathcal{F}_{axis}(e_j, A) \right\},
\end{align}
where $\mathcal{F}_{axis}(e_j, A)$ is an indicator function that determines whether element $e_j$ overlaps with axis $A$. To avoid duplicate comparisons and to emphasize elements with higher uniqueness, we define the notion of a \emph{race group} for each element $e_i$. Specifically, we traverse $G_i$ to identify all elements that share the same class and tag name; these elements form $e_i$'s race group, capturing repeated items such as list entries or navigation links. The Group Count for the web page is then computed using Algorithm~\ref{a2}. To weight elements according to uniqueness, we define $e_i$'s race weight $w_i$ as $\frac{1}{|e_i.\text{raceGroup}| \cdot C}$. This formulation ensures that elements with high repetition receive a lower weight, while unique elements contribute more to the overall evaluation.
Finally, the GDA score is computed by comparing the number of elements in corresponding generated and target groups. For a pair of groups $(S_j, T_j)$, the score is defined as:
\begin{equation}
\text{score}(S_j, T_j) =
\begin{cases}
1 \cdot w_j, & \text{if } |S_j| = |T_j| \\
0, & \text{otherwise}
\end{cases}.
\end{equation}
The overall GDA score for the page is then obtained by summing over all group comparisons.

\begin{algorithm}[h]
\caption{Compute Group Count}\label{a2}
\KwIn{$E$ : set of all target visible elements}
\KwOut{Group Count $C$ }

$V \gets \emptyset$ \tcp*{set of viewed elements}
$C \gets 0$ \tcp*{initialize count}

\ForEach{$e \in E$}{
    \If{$e \notin V$}{
        $C \gets C + 1$\;
        $V \gets V \cup \{e\}$\;
        \ForEach{$r \in e.\text{raceGroup}$}{
            $V \gets V \cup \{r\}$\;
        }
    }
}
\Return $C$\;
\end{algorithm}

\subsubsection*{\textbf{Style Difference of Associated Elements (SDA)}} In addition to layout consistency, we evaluate fine-grained style consistency for key visual attributes, such as foreground color, background color, font size, and border radius. For each associated element pair $(s_i, t_i)$, we compute a score for each style attribute and average these scores across all attributes. Finally, the per-element scores are weighted by their race weight $w_i$ to calculate the overall SDA score between the generated and reference pages.

\subsubsection*{\textbf{Training Implementation.}} Taking advantage of real-time evaluation, we directly use the computed scores as rewards to optimize the model's generation ability. To this end, we develop ALISA, which leverages a Web Server Tool to asynchronously compute reward scores for policy model rollouts. Specifically, a webpage screenshot and a query prompt are provided as input to the ALISA policy model. During rollouts, the model generates $N$ candidate responses for each sample $x_i$. Each response is rendered using a WebDriver and then compared with the target webpage to obtain its RDA, GDA, and SDA scores. The final reward for each sample is computed as:
\begin{equation}
R_{i,j} = \alpha \text{RDA}_j + \beta \text{GDA}_j + \gamma \text{SDA}_j,
\end{equation}
where $\alpha, \beta, \gamma$ are weighting coefficients. The advantage $A_{i,j}$ for each sample is computed by normalizing its reward through mean–standard deviation scaling: $A_{i,j}=\frac{R_{i,j}-\mathrm{mean}(\mathrm{R})}{\mathrm{std}(\mathrm{R})}$. Finally, the policy is optimized based on GRPO:
\begin{multline}
L_{\text{policy}} = \frac{1}{N} \sum_{j=1}^{N} \min \Big( \rho_i A_j, \operatorname{clip}(\rho_j, 1-\epsilon, 1+\epsilon) A_j \Big)\\ -\lambda\mathbb{D}_{KL}\left[\pi_{\theta}||\pi_{r e f}\right],
\end{multline}
where $\rho_i = \dfrac{\pi_\theta(a_i|s_i)}{\pi_{ref}(a_i|s_i)}$, $\pi_\theta(\cdot)$ denotes the policy function, $\lambda$ is a hyperparameter, $a_i$ represents the probability of the next-token action, and $\mathbb{D}_{KL}$ denotes the KL divergence penalty.

\section{Experiments}
\subsection{Experimental Setup}
In our experiments, we evaluate the WebUI-to-Code inference performance of WebRenderBench across VLMs of varying scales and architectures. Furthermore, we conduct reinforcement learning experiments based on ALISA to validate its effectiveness. Our analysis centers on examining whether current VLMs can meet user expectations with respect to layout and style consistency.

\subsubsection*{\textbf{Metrics.}} We adopt the RDA, GDA, and SDA metrics introduced in Section~\ref{s33} to evaluate the WebUI-to-Code reasoning and training performance of VLMs under different Group Count ranges. In addition, our training experiments further compare traditional metrics such as Jaccard similarity and CLIP scores, as well as the WUB metric from Web2Code, to comprehensively validate the effectiveness of our approach.

\subsubsection*{\textbf{Models.}} We conduct experiments on seventeen VLMs, which include four closed-source models: GPT-4.1-mini \cite{openai2024gpt4technicalreport}, ByteDance-Seed-1.6-Vision \cite{bytedance_seed1_6}, Qwen-VL-Plus \cite{qwen2.5-VL}, Kimi-0905 \cite{kimiteam2025kimik15scalingreinforcement}, and GLM-4V-Plus-0111 \cite{vteam2025glm45vglm41vthinkingversatilemultimodal}. The open-source models comprise the Llama family \cite{grattafiori2024llama3herdmodels}, the Qwen2.5-VL family \cite{qwen2.5-VL}, the DeepSeek-VL2 family \cite{wu2024deepseekvl2mixtureofexpertsvisionlanguagemodels}, and the InternVL3 family \cite{wang2025internvl3_5}. These models span parameter scales ranging from 3B to 90B, ensuring comprehensive coverage.

\subsubsection*{\textbf{Settings.}} During inference, all open-source models are run using vLLM with the temperature set to zero to ensure consistent outputs. Inference experiments are conducted on the WebRenderBench test set using eight NVIDIA H20 GPUs. For ALISA training, to balance time and computational cost, we sample 4,000 training instances with screenshot heights below 3,000 pixels. The number of rollout samples $N$ is set to 3, the hyperparameters $\alpha, \beta, \gamma$ are set to 0.6, 0.2, and 0.2, respectively, and the KL divergence coefficient $\lambda$ is set to the conventional value of 0.01. The batch size is set to 128, and the learning rate is 1e-6.

\subsection{Main Results}

\begin{table*}[h]
\caption{Comparison of RDA, GDA, and SDA scores (\%) across different VLMs and Group Count ranges. The best and second-best scores in each category are highlighted in \textbf{bold} and \uline{underlined}, respectively.}
\label{t2}
\resizebox{\textwidth}{!}{%
\begin{tabular}{@{}p{5.5cm}|ccc|ccc|ccc|ccc|ccc|ccc@{}}
\toprule
{\multirow{2}{*}{\textbf{Model}}} & \multicolumn{3}{c|}{\textbf{0-50}}         & \multicolumn{3}{c|}{\textbf{50-100}}       & \multicolumn{3}{c|}{\textbf{100-150}}     & \multicolumn{3}{c|}{\textbf{150-200}}     & \multicolumn{3}{c|}{\textbf{200-400}}     & \multicolumn{3}{c}{\textbf{400+}} \\
{}                                & RDA   & GDA   & {SDA}   & RDA   & GDA   & {SDA}   & RDA  & GDA   & {SDA}   & RDA  & GDA   & {SDA}   & RDA  & GDA   & {SDA}   & RDA       & GDA       & SDA       \\ \midrule
\multicolumn{19}{c}{Close   Source VLMs}                                                                                                                                                                                                                                                                              \\ \midrule
GPT-4.1-mini-20250414           & \textbf{28.62} & {\ul 48.83}    & \textbf{50.40} & \textbf{13.44} & 31.40          & \textbf{36.56} & \textbf{8.47} & \textbf{29.56} & \textbf{32.96} & \textbf{8.08} & \textbf{31.68} & \textbf{31.60} & \textbf{4.83} & \textbf{22.93} & \textbf{25.31} & \textbf{5.95} & \textbf{18.33} & \textbf{24.13} \\
ByteDance-Seed-1.6-Vision & 18.94          & 43.31          & 39.72          & 10.24          & {\ul 33.38}    & 31.51          & 5.36          & 22.60          & 23.61          & 5.47          & 22.41          & 23.38          & 3.12          & 15.23          & 17.23          & 2.78          & 11.48          & 15.23          \\
Qwen-VL-Plus           & {\ul 19.12}    & \textbf{49.90} & {\ul 49.48}    & {\ul 11.51}    & \textbf{40.36} & {\ul 35.09}    & 6.13          & 23.08          & 27.48          & {\ul 5.72}    & {\ul 26.83}    & 26.30          & {\ul 4.03}    & {\ul 20.70}    & {\ul 22.54}    & {\ul 3.98}    & {\ul 15.40}    & {\ul 18.94} \\
Kimi-0905        & 14.43 & 39.30 & 39.89 & 7.19 & 23.92 & 18.83 & 3.75 & 16.10 & 14.14 & 3.00 & 13.40 & 10.31 & 2.43 & 11.80 & 8.41 & 1.06 & 8.12 & 4.43 \\
GLM-4V-Plus-0111 & 16.75 & 45.73 & 41.54 & 4.05 & 22.13 & 16.89 & 4.02 & 12.73 & 12.06 & 2.91 & 17.15 & 9.88  & 1.87 & 10.00 & 6.64 & 0.78 & 6.37 & 3.41 \\ \midrule
\multicolumn{19}{c}{Open   Source   VLMs}                                                                                                                                                                                                                                                                             \\ \midrule
{Llama-3.2-90B-Vision-Instruct}   & 15.54 & 37.06 & 30.84 & 3.31 & 17.83 & 10.14 & 1.77 & 12.05 & 6.67 & 1.01 & 8.79 & 4.86 & 0.62 & 6.23 & 3.28 & 0.38 & 4.66 & 2.29      \\
{InternVL3-78B}    & 11.54 & 32.74 & 33.14 & 6.35 & 25.69 & 22.39 & 4.61 & 21.15 & 19.18 & 4.25 & 20.39 & 17.93 & 2.64 & 15.8 & 14.09 & 1.87 & 12.14 & 10.65          \\
{Qwen2.5-VL-72B-Instruct}         & 17.22 & 43.55 & {43.46} & 8.44  & 31.00 & {28.48} & 5.50 & 25.27 & {24.50} & 4.12 & 22.33 & {22.38} & 2.90 & 17.69 & {16.94} & 1.62      & 11.76     & 11.90     \\
{Qwen2.5-VL-32B-Instruct}         & 18.90 & 47.41 & {45.93} & 9.87  & 32.77 & {31.35} & {\ul 6.78} & {\ul 27.91} & {\ul 27.67} & 5.27 & 25.44 & {\ul 26.48} & 3.78 & 20.25 & {20.86} & 2.07      & 12.92     & 16.46     \\
{Deepseek-VL2 (28B)}              & 10.61 & 36.11 & {30.85} & 4.90  & 23.22 & {16.87} & 3.07 & 18.00 & {13.27} & 2.08 & 15.15 & {10.93} & 1.29 & 11.33 & {7.81}  & 0.65      & 7.44      & 5.16      \\
{Deepseek-VL2-Small (16B)}        & 7.81  & 32.00 & {22.52} & 3.29  & 22.55 & {11.49} & 2.15 & 17.74 & {8.82}  & 1.56 & 14.94 & {7.21}  & 0.94 & 11.45 & {4.95}  & 0.54      & 7.98      & 3.54      \\
{InternVL3-14B}                   & 9.49  & 30.09 & {29.30} & 4.53  & 19.96 & {16.37} & 2.90 & 15.79 & {13.25} & 2.13 & 13.11 & {11.59} & 1.48 & 10.6  & {9.01}  & 0.99      & 7.86      & 6.98      \\
{Llama-3.2-11B-Vision-Instruct}   & 5.32 & 29.24 & 14.30 & 2.94 & 21.04 & 11.30 & 1.54 & 17.50 & 7.90 & 0.94 & 13.34 & 5.82 & 0.56 & 9.36 & 4.34 & 0.32 & 6.78 & 3.22      \\
{InternVL3-8B}                    & 8.50  & 30.75 & {30.11} & 5.07  & 22.84 & {20.28} & 3.44 & 18.76 & {17.08} & 2.52 & 15.84 & {15.06} & 1.79 & 13.17 & {11.95} & 1.21      & 9.79      & 9.81      \\
{Qwen2.5-VL-7B-Instruct}          & 13.13 & 40.10  & {37.09} & 6.95  & 29.88 & {24.96} & 4.97 & 25.17 & {21.72} & 3.95 & 22.71 & {19.83} & 2.83 & 18.62 & {15.67} & 1.75      & 12.98     & 12.23     \\
{Qwen2.5-VL-3B-Instruct}          & 8.42  & 27.39 & {22.08} & 3.83  & 21.85 & {14.49} & 2.75 & 18.61 & {12.86} & 2.25 & 16.97 & {12.21} & 1.49 & 13.60  & {8.98}  & 0.90       & 9.97      & 6.78      \\
{Deepseek-VL2-Tiny (3B)}          & 2.83  & 26.68 & {8.52}  & 0.94  & 16.37 & {4.28}  & 0.54 & 12.47 & {2.99}  & 0.35 & 9.42  & {2.33}  & 0.19 & 7.26  & {1.59}  & 0.11      & 5.07      & 1.10      \\ \bottomrule
\end{tabular}%
}
\end{table*}

\subsubsection*{\textbf{Inference Results.}} Table~\ref{t2} shows the inference performance of the models across Group Count ranges from 0 to over 400, allowing us to explore the limits of model performance. Overall, GPT-4.1-mini-20250414 and Qwen-VL-Plus achieve the highest average scores, and performance generally improves with increasing model size. The results indicate that all models demonstrate better alignment in style and group consistency; however, RDA declines sharply when the Group Count exceeds 50, suggesting that models can effectively align corresponding elements only on simpler webpages and highlighting the remaining challenges in the WebUI-to-Code task. In addition, while GDA and SDA exhibit similar overall proportions, their ratios relative to RDA differ by a factor of two to three. This finding guides the setting of reward weight ratios in our framework.

\begin{table*}[h]
\caption{Comparison of RDA, GDA, and SDA scores (\%) under different training strategies across Group Count ranges. The best and second-best scores in each category are highlighted in \textbf{bold} and \uline{underlined}, respectively.}
\label{t3}
\resizebox{\textwidth}{!}{%
\begin{tabular}{@{}l|l|ccc|ccc|ccc|ccc|ccc|ccc@{}}
\toprule
\multirow{2}{*}{\textbf{Model}}      & \multirow{2}{*}{\textbf{Method}} & \multicolumn{3}{c|}{\textbf{0-50}}               & \multicolumn{3}{c|}{\textbf{50-100}}            & \multicolumn{3}{c|}{\textbf{100-150}}           & \multicolumn{3}{c|}{\textbf{150-200}}           & \multicolumn{3}{c|}{\textbf{200-400}}           & \multicolumn{3}{c}{\textbf{400+}}               \\
                                     &                                  & RDA            & GDA            & SDA            & RDA           & GDA            & SDA            & RDA           & GDA            & SDA            & RDA           & GDA            & SDA            & RDA           & GDA            & SDA            & RDA           & GDA            & SDA            \\ \midrule
Qwen2.5-VL-3B-Instruct               & +SFT                             & 4.13           & 11.83          & 11.58          & 2.07          & 13.60          & 10.08          & 1.53          & 12.85          & 9.40           & 1.38          & 12.70           & 9.56           & 0.95          & 9.92           & 7.61           & 0.75          & 7.18           & 7.80           \\
Qwen2.5-VL-7B-Instruct               & +SFT                             & 6.08           & 12.37          & 12.36          & 5.21          & 13.77          & 11.69          & 2.01          & 12.87          & 9.73           & 3.26          & 12.16          & 11.03          & 1.01          & 9.96           & 8.82           & 0.73          & 8.28           & 8.92           \\ \midrule
\multirow{5}{*}{ALISA-Qwen2.5-VL-3B} & Default                          & {\ul 33.09}    & {\ul 60.05}    & \textbf{52.84} & 8.17          & \textbf{38.69} & {\ul 29.99}    & 5.63          & {\ul 35.23}    & {\ul 27.27}    & 4.34          & {\ul 33.49}    & {\ul 26.07}    & \textbf{3.08} & {\ul 27.44}    & {\ul 20.86}    & 1.98          & \textbf{19.78} & 16.33          \\
                                     & w/ RDA + GDA                            & 27.88          & 57.19          & 45.13          & 7.68          & 34.98          & 25.13          & 5.40          & 31.91          & 22.83          & 3.88          & 29.82          & 21.49          & 2.88          & 24.55          & 16.87          & 1.58          & 16.98          & 11.41          \\
                                     & w/ RDA                            & 27.12          & 52.37          & {42.19}          & 7.69          & 34.14          & {24.79}          & 5.42          & 30.58          & {22.11}          & 3.62          & 28.33          & {21.26}          & 2.98          & 24.40          & {17.71}          & 1.92          & 17.27          & 13.51          \\
                                     & w/ GDA                            & 23.93          & 49.21          & 42.07          & 6.57          & 33.71          & 25.25          & 4.32          & 29.84          & 22.47          & 3.60          & 28.67          & 21.88          & 2.39          & 24.10          & 18.36          & 1.73          & 17.77          & 14.73          \\
                                     & w/ SDA                            & 16.03          & 42.61          & 42.26          & 6.34          & 27.15          & 24.21          & 4.37          & 23.24          & 21.43          & 3.13          & 20.53          & 19.27          & 2.17          & 16.04          & 14.91          & 1.17          & 10.46          & 11.06          \\
                                     & w Jaccard                        & 15.91          & 42.53          & 39.30          & 6.23          & 26.36          & 27.49          & 4.12          & 24.31          & 25.63          & 3.14          & 23.83          & 25.79          & 2.08          & 19.33          & 19.86          & 1.42          & 14.16          & 16.30          \\ \midrule
\multirow{5}{*}{ALISA-Qwen2.5-VL-7B} & Default                          & \textbf{33.55} & \textbf{60.28} & {\ul 52.22}    & \textbf{8.97} & {\ul 38.66}    & \textbf{30.88} & \textbf{5.97} & \textbf{36.13} & \textbf{28.61} & \textbf{4.99} & \textbf{35.26} & \textbf{29.47} & {\ul 3.07}    & \textbf{28.23} & \textbf{21.79} & \textbf{2.40} & {\ul 19.72}    & \textbf{17.28} \\
                                     & w/ RDA + GDA                            & 28.00          & 57.21          & 46.93          & {\ul 8.96}    & 32.33          & 29.39          & 5.83    & 30.25          & 25.00          & {\ul 4.40}    & 29.38          & 23.23          & 3.05          & 24.96          & 18.83          & {\ul 2.30}    & 17.03          & 15.58          \\
                                     & w/ RDA                            & 28.76          & 56.19          & {45.12}          & 8.33          & 31.13          & {29.18}          & {\ul 5.92}          & 29.79          & {25.06}          & 4.18          & 29.64          & {23.10}          & 3.05          & 23.99          & {17.92}          & 2.28          & 18.75          & 16.28    \\
                                     & w/ GDA                            & 26.13          & 54.02          & 45.45          & 7.98          & 31.58          & 29.14          & 5.10          & 29.68          & 25.22          & 4.13          & 28.94          & 23.51          & 2.96          & 19.12          & 17.85          & 2.24          & 16.83          & {\ul 16.52}    \\
                                     & w/ SDA                            & 23.25          & 50.60          & 45.10          & 7.59          & 30.78          & 29.35          & 5.69          & 26.90          & 25.11          & 4.29          & 24.28          & 23.06          & 2.96          & 19.13          & 18.57          & 2.27          & 13.30          & 16.13          \\
                                     & w Jaccard                        & 20.92          & 48.12          & 44.38          & 7.50          & 29.51          & 26.57          & 5.07          & 24.58          & 23.19          & 3.74          & 21.97          & 21.09          & 2.60          & 17.36          & 16.69          & 1.63          & 11.73          & 13.09          \\ \bottomrule
\end{tabular}%
}
\end{table*}

\subsubsection*{\textbf{Training Results.}} To evaluate the feasibility and effectiveness of our proposed ALISA framework in optimizing the WebUI-to-Code task on asymmetric code, we sampled a subset of the training data and conducted training experiments, assessing model performance using the same setup as the inference experiments. We use Qwen2.5-VL-Instruct as the backbone model for further fine-tuning and compare results between SFT using the raw crawled target webpage code as labels and reinforcement learning with Jaccard similarity based on temporal sequence alignment as a reward. The results, presented in Table~\ref{t3}, show that SFT with either the 3B or 7B model leads to performance degradation compared to the vanilla Qwen2.5-VL, demonstrating that directly using noisy source code as labels is ineffective. In contrast, models fine-tuned with ALISA achieve substantial improvements across all metrics. Specifically, ALISA-Qwen2.5-VL-7B surpasses GPT-4.1-mini-20250414 by 4.93\%, 11.45\%, and 1.82\% in RDA, GDA, and SDA, respectively, within the Group Count range [0, 50), highlighting significant improvements in layout and style consistency. Across other Group Count ranges, it also outperforms Qwen2.5-VL-72B-Instruct, further demonstrating the effectiveness of our framework in optimizing model performance with asymmetric code. Additionally, the ``w Jaccard'' experiments show performance gains over the vanilla model across all metrics, indicating that Jaccard similarity can serve as a naive method to partially mitigate noise in the source code. However, the improvements are smaller than those achieved with the default ALISA approach, reflecting its limitations in goal-directed optimization.

\subsection{Analysis and Discussion}
\subsubsection*{\textbf{Ablation Study}} We further present ablation experiments in Table~\ref{t3} to examine the effect of using only one of the RDA, GDA, or SDA scores during training. The results indicate that retaining any single metric improves performance compared with the vanilla model. However, when relying solely on style consistency (w/ SDA), the model performs significantly worse than when focusing on layout consistency, and in some cases even falls below the results of ``w Jaccard'' training. This finding suggests that layout consistency should be prioritized to enhance WebUI-to-Code performance. Moreover, ``w/ RDA'' yields substantial improvements in both RDA and GDA, while ``w/ GDA'' primarily improves GDA. Overall, compared with w/ RDA alone, incorporating style consistency alongside layout consistency produces additional gains in both layout and style consistency. We further analyze the impact of the $\alpha$, $\beta$, and $\gamma$ weights on model performance in Appendix~\ref{a_reward}.

\begin{table}[h]
\caption{Comparison of WUB (\%) across different training strategies, with parentheses indicating results obtained using different datasets. The best and second-best scores in each category are highlighted in \textbf{bold} and \uline{underlined}, respectively.}
\label{t4}
\resizebox{\columnwidth}{!}{%
\begin{tabular}{lp{1.5cm}}
\toprule
\textbf{Model}                                                                                & \textbf{WUB (\%)}   \\ \midrule
Qwen2.5-VL-3B-Instruct                                                               & 72.38 \\
Qwen2.5-VL-7B-Instruct                                                               & 75.94 \\ \midrule
Qwen2.5-VL-3B-Instruct-SFT   (Web2Code)                                              & 64.60 \\
Qwen2.5-VL-3B-Instruct-SFT (WebRenderBench)                                          & 59.58 \\ \midrule
ALISA-Qwen2.5-VL-3B (Web2Code)       & 74.16 \\
ALISA-Qwen2.5-VL-3B (WebRenderBench) & 74.98 \\
ALISA-Qwen2.5-VL-7B (Web2Code)       & {\ul 76.01} \\
ALISA-Qwen2.5-VL-7B (WebRenderBench) & \textbf{76.47} \\ \bottomrule
\end{tabular}%
}
\end{table}

\subsubsection*{\textbf{Evaluation on WUB}} To further verify the effectiveness of ALISA in improving WebUI-to-Code performance and to assess the influence of webpage quality on training, we conduct evaluations using the Webpage Understanding Benchmark (WUB) from Web2Code. We compare models trained with ALISA against both vanilla Qwen2.5-VL-Instruct inference and SFT. For consistency, Qwen-VL-Plus is used as the question–answering model in all WUB evaluations. As shown in Table~\ref{t4}, ALISA-trained models achieve gains of 2.6\% and 0.12\% over vanilla Qwen2.5-VL-3B-Instruct and Qwen2.5-VL-7B-Instruct, respectively. In addition, we randomly sample 4,000 instances from the Web2Code dataset, matching the scale used in ALISA training, and conduct both SFT and reinforcement learning experiments. The results indicate that ALISA with Web2Code also improves performance, although the gains are smaller than those obtained with WebRenderBench. Furthermore, using a small subset of Web2Code for SFT results in performance degradation, indicating that SFT is more effective when trained on a larger dataset. Training with WebRenderBench using noisy HTML data as labels results in even greater degradation, highlighting the practicality of the ALISA framework. Finally, although baseline WUB scores are already relatively high, the improvements achieved by ALISA are less pronounced, likely due to the lower complexity of webpages in WUB and its limited capacity to fully capture layout and style consistency.

\begin{table}[h]
\caption{Comparison of Jaccard similarity and CLIP scores across different Group Count ranges. The highest and second-highest scores in each category are shown in \textbf{bold} and \uline{underlined}, respectively.}
\label{t5}
\resizebox{\columnwidth}{!}{%
\begin{tabular}{@{}l|cc|cc|cc@{}}
\toprule
\multicolumn{1}{c|}{\multirow{2}{*}{\textbf{Model}}} & \multicolumn{2}{c|}{\textbf{0-200}} & \multicolumn{2}{c|}{\textbf{200-400}} & \multicolumn{2}{c}{\textbf{400+}} \\
\multicolumn{1}{c|}{}                                & Jaccard           & CLIP            & Jaccard            & CLIP             & Jaccard          & CLIP           \\ \midrule
Qwen2.5-VL-3B-Instruct                               & 32.80             & 69.99           & 20.54              & 68.90            & 15.44            & 69.02          \\
Qwen2.5-VL-72B-Instruct                              & {\ul 45.25}             & \textbf{74.78}           & 28.91              & \textbf{74.19}            & 20.98            & \textbf{72.86}          \\
ALISA-Qwen2.5-VL-3B                                  & 44.82             & 72.35           & {\ul 31.72}              & 72.55            & {\ul 24.10}            & 71.76          \\
ALISA-Qwen2.5-VL-3B (w Jaccard)                      & \textbf{51.36}             & 72.33           & \textbf{35.69}              & 71.85            & \textbf{27.30}            & 70.39          \\ \bottomrule
\end{tabular}%
}
\end{table}

\subsubsection*{\textbf{Comparison on Jaccard and CLIP}} To further evaluate our method using both text-based and vision-based metrics, Table~\ref{t5} presents a comparison with baseline models in terms of Jaccard similarity and CLIP scores. The CLIP scores are computed using the pre-trained model ``clip-ViT-B-32''\footnote{https://huggingface.co/openai/clip-vit-base-patch32}. The results show that our method consistently outperforms vanilla Qwen2.5-VL-3B-Instruct on both metrics. Among all methods, the model trained with Jaccard similarity as the reward achieves the highest Jaccard score, while Qwen2.5-VL-72B-Instruct shows a clear advantage in CLIP score. Together with the results in Table~\ref{t2}, these findings suggest that larger models are more likely to generate outputs whose overall visual quality is closer to the target webpages, although discrepancies remain at finer-grained levels, leading to the observed gap with the SDA score. This indicates that both text-based and vision-based evaluations have inherent limitations.

\begin{figure}[ht]
  \includegraphics[width=0.8\columnwidth]{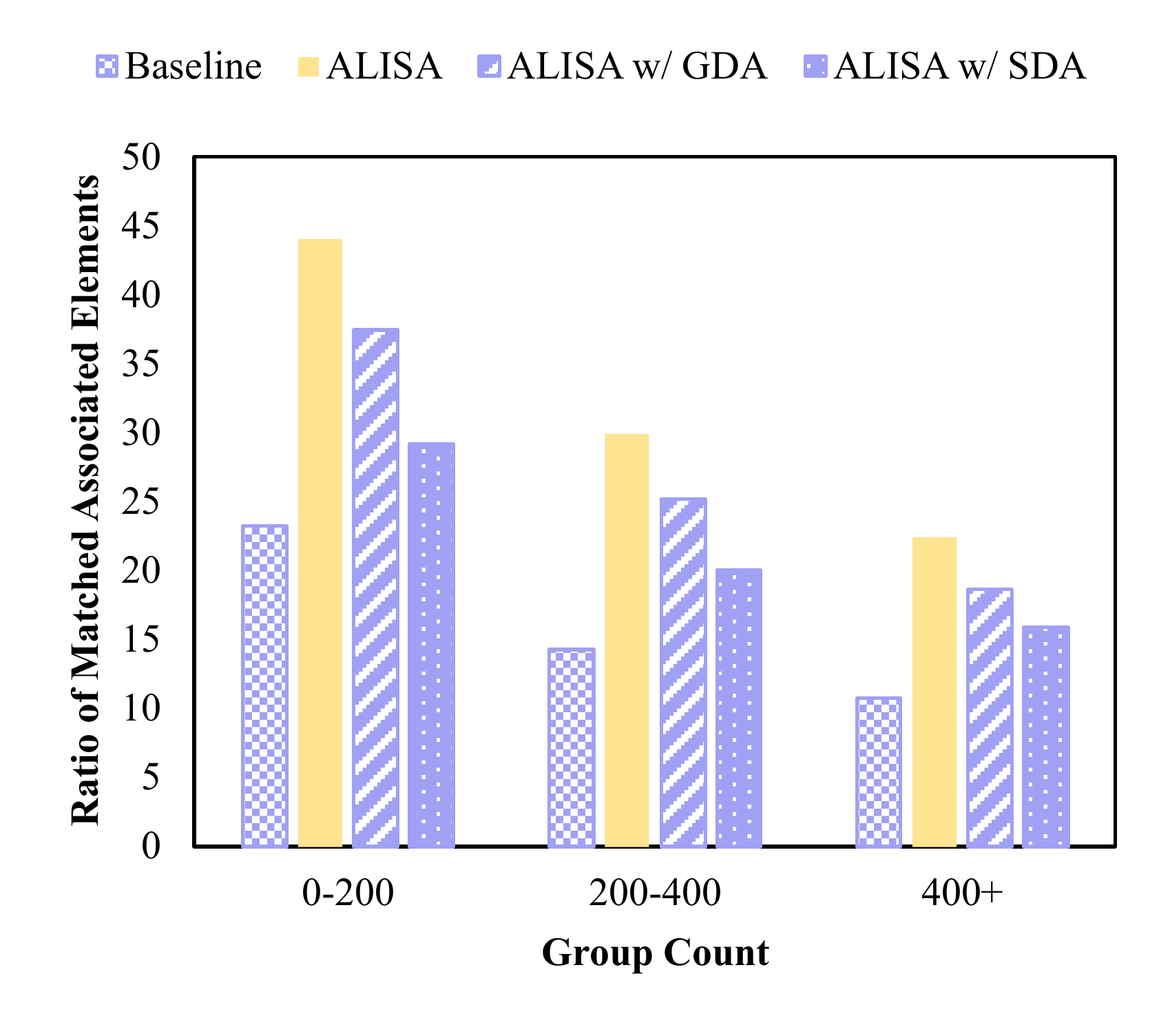}
  \caption{Ratio of Matched Associated Elements across different Group Count ranges.}
  \label{f4}
\end{figure}

\subsubsection*{\textbf{Effectiveness Analysis of RDA}} To further examine the generally low/ RDA scores observed in the main experiments and to understand why improvements in RDA also benefit other metrics, we analyze the proportion of associated elements in pages generated by Qwen2.5-VL-3B-Instruct that can be successfully matched to their target pages. We compare these results with those obtained after optimizing the model using our ALISA framework, as shown in Figure~\ref{f4}. In pages generated by the vanilla model, fewer than 25\% of elements are successfully matched to associated elements in the target webpages, which largely explains the low/ RDA scores. After training with RDA-targeted reward optimization, the proportion of matched associated elements roughly doubles compared to the baseline. This increase likely contributes to improved group consistency, indirectly enhancing GDA as well.

\section{Conclusion}

In this paper, we present WebRenderBench, a large-scale and diverse benchmark for WebUI-to-Code, accompanied by a novel code-level metric that evaluates layout and style consistency based on rendered webpages. To further enhance generation quality, we introduce ALISA, an automated layout and style inspection agent that leverages this metric as a reinforcement learning reward. Experimental results demonstrate that integrating ALISA substantially improves the performance of VLMs on complex, real-world webpages. WebRenderBench offers a robust platform for benchmarking and advancing research in WebUI-to-Code.


\bibliographystyle{ACM-Reference-Format}
\bibliography{sample-base}

\appendix

\section{Ethical Statement}

The benchmark constructed in this work is derived from real company portal websites. Due to cross-origin resource restrictions, some website images have been replaced, and certain text data have been modified or rewritten with LLMs. Personal or sensitive information would be removed when the dataset is released to protect privacy. As a result, the dataset may deviate from the original webpages. All data collection and processing comply with applicable laws and regulations, and the dataset is intended solely for academic research purposes without any commercial use. 

\section{Prompt Example}

\begin{figure}[h]
  \includegraphics[width=\columnwidth]{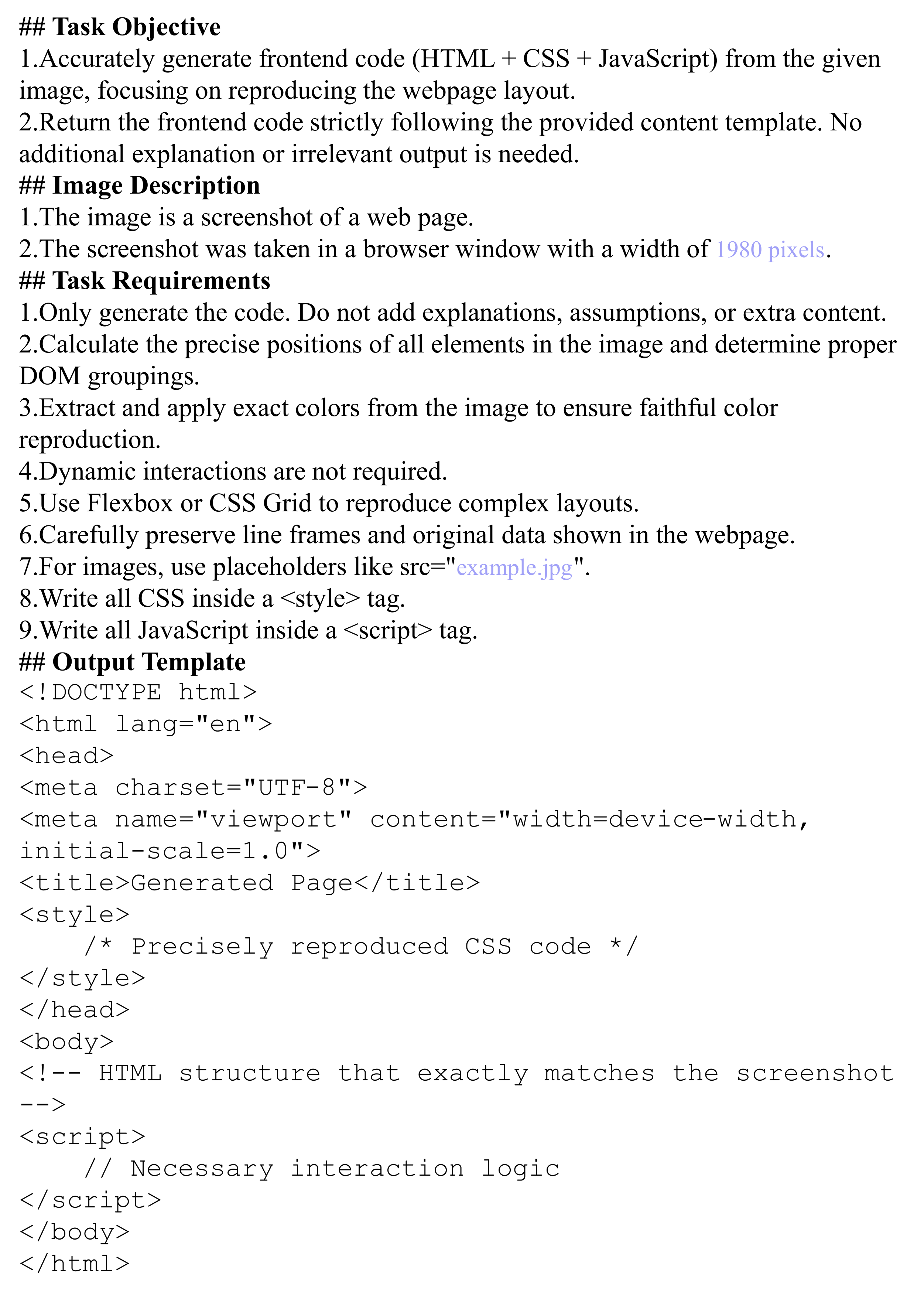}
  \caption{Example prompts for VLM inference and training, where the purple sections indicate editable variables. Additionally, an \texttt{<image>} placeholder is prepended at the beginning of the input.}
  \label{a1}
\end{figure}

Figure~\ref{a1} presents example prompts used for VLM inference and training. To ensure that the model generates code as close to the original as possible, the prompts instruct it to use only native CSS styles for layout. Additionally, all generated image resources are replaced with ``example.jpg'' during generation, and relevant resources are later substituted using associated elements.

\section{Reward Weight Analysis}
\label{a_reward}

\begin{figure*}[h]
  \includegraphics[width=\textwidth]{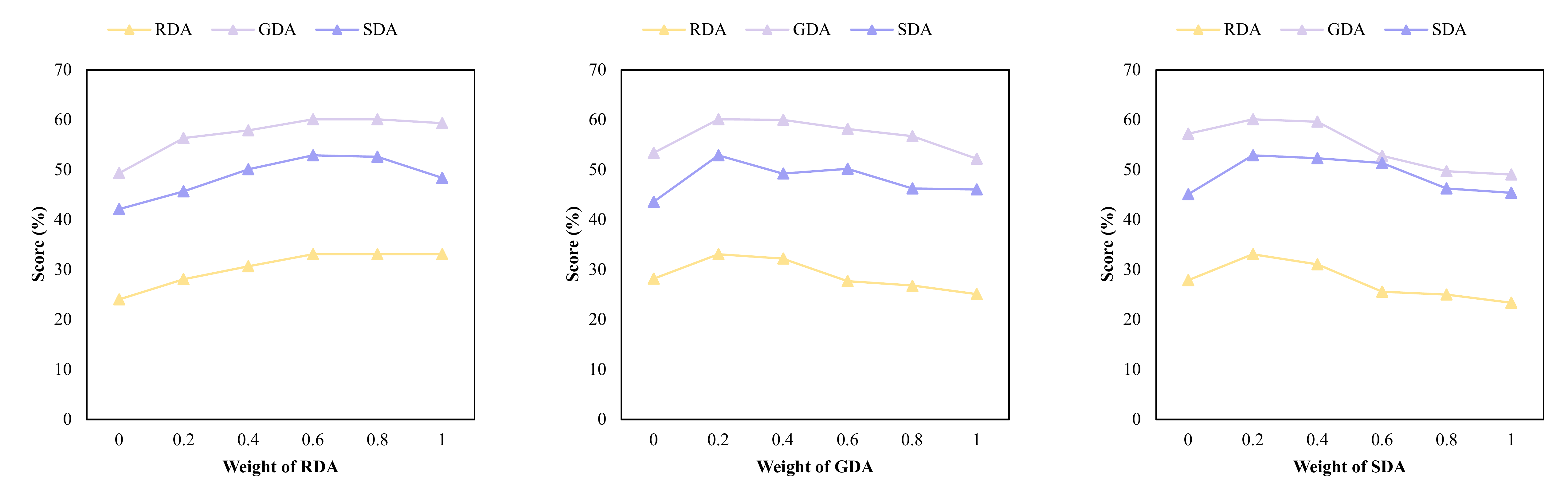}
  \caption{Impact of RDA, GDA, and SDA reward weights on model performance using Qwen2.5-VL-3B-Instruct as the backbone model.}
  \label{a2}
\end{figure*}

We further investigate how varying the reward weights of RDA, GDA, and SDA affects model performance, as illustrated in Figure~\ref{a2}. In this analysis, only one weight parameter is adjusted at a time while the other two remain fixed at their default values. The results indicate that increasing the RDA weight consistently improves all evaluation metrics. However, when the RDA weight is set to 1, the relatively small proportion of SDA leads to a noticeable drop in SDA performance. Moreover, variations in the GDA and SDA weights have the most pronounced effect on SDA. In particular, setting the SDA weight too low or too high causes performance degradation. This is likely because an excessively large SDA weight shifts the model’s attention away from layout consistency during training, which in turn reduces the overall quality of the generated webpages.

\begin{figure*}[h]
  \includegraphics[width=\textwidth]{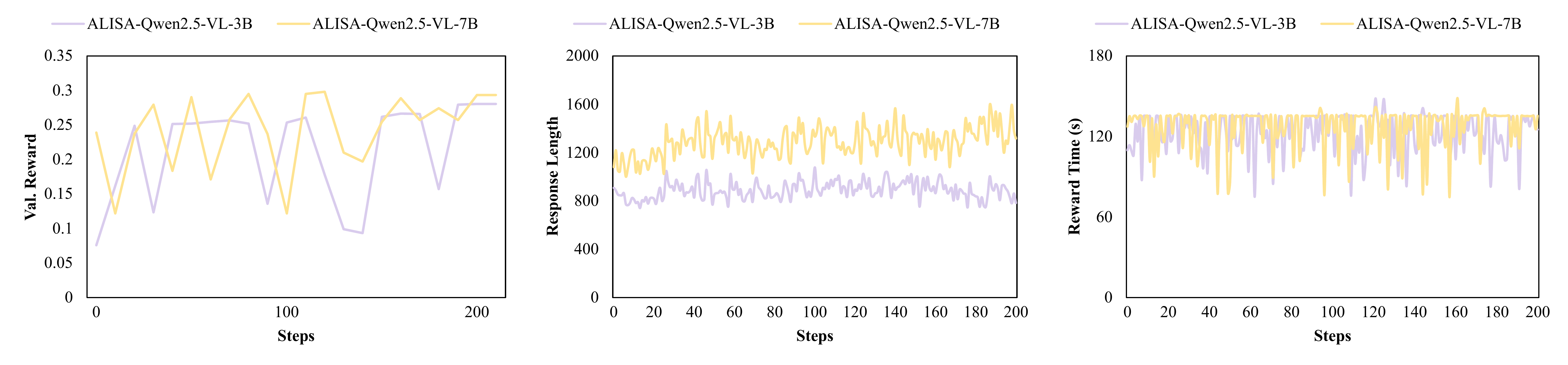}
  \caption{Comparison of webpage generation consistency between the ALISA framework and other Qwen2.5-VL-Instruct models, relative to the ground truth.}
  \label{a3}
\end{figure*}

\section{Training Process Analysis}

Figure~\ref{a3} presents the changes in validation reward scores, rollout response lengths, and reward computation times over the first 200 training iterations for ALISA-Qwen2.5-VL-3B and ALISA-Qwen2.5-VL-7B. We observe that the reward scores of both models fluctuate during training, which may be attributed to higher sample complexity in certain intervals, resulting in lower-quality rollouts due to the limited capacity of the pre-trained models. Nevertheless, both models exhibit an overall upward trend in validation reward scores.

In terms of response length, both models show a slight increase as training progresses, with the 7B model consistently producing longer outputs than the 3B model. Further examination of the generated results reveals that larger models tend to provide more detailed CSS code, which is also evident in the case studies. Regarding computation time, the average reward calculation remains around two minutes, with minor fluctuations. To support efficient parallel processing, we deployed the ALISA web server with 64 workers, enabling simultaneous batch computation and minimizing latency. Overall, this response time meets the requirements for online inference.

\section{Human Evaluation}

\begin{table}[h]
\caption{Results of human evaluation. $\kappa$ denotes the inter-rater agreement score. The highest and second-highest values in each category are highlighted in bold and \uline{underlined}, respectively.}
\label{t6}
\resizebox{\columnwidth}{!}{%
\begin{tabular}{@{}lccccc@{}}
\toprule
\multicolumn{1}{c}{\textbf{Model}} & \textbf{Layout} & \textbf{Style} & \textbf{Content} & \textbf{Avg.} & \textbf{$\kappa$ (\%)} \\ \midrule
ALISA-Qwen2.5-VL-3B                & \textbf{5.60}            & \textbf{5.13}           & \textbf{6.50}             & \textbf{5.74} & 71.42          \\
ALISA-Qwen2.5-VL-3B   (w/ SDA)     & 4.53            & 4.33           & {\ul 5.32}             & 4.73          & 68.71          \\
Qwen2.5-VL-3B-Instruct             & 4.22            & 4.52           & 4.32             & 4.35          & 81.22          \\
Qwen2.5-VL-7B-Instruct             & 4.28            & 4.42           & 4.48             & 4.39          & 74.15          \\
Qwen2.5-VL-32B-Instruct            & 4.72            & 4.95           & 5.02             & 4.90          & 71.79          \\
Qwen2.5-VL-72B-Instruct            & {\ul 4.78}            & {\ul 5.03}           & 5.22             & {\ul 5.01}    & 82.87          \\ \bottomrule
\end{tabular}%
}
\end{table}

To further verify the quality of the generated webpages and the reliability of our evaluation metrics, we conducted a human evaluation involving three Vue developers, each with over three years of experience in front-end development. The evaluators rated each generated webpage on a scale from 1 (poor) to 10 (perfectly consistent) across three dimensions:

\begin{itemize}
    \item Layout Consistency: The extent to which the overall structure and relative positioning of elements match the reference screenshot.
    \item Style Consistency: The similarity of visual attributes such as colors, fonts, spacing, borders, and shadows to the reference.
    \item Content Accuracy: The accuracy of reproduced textual, image, and icon content.
\end{itemize}

The results are summarized in Table~\ref{t6}. We also calculated Cohen's kappa ($\kappa$) coefficient to assess inter-rater agreement, which exceeded 70\% across all models, indicating a high level of consistency among evaluators. As shown, ALISA-Qwen2.5-VL-3B achieves notably higher scores in layout consistency and content accuracy compared with other models, which aligns well with our automatic evaluation results. Furthermore, the average scores across all models are around the mid-range (approximately 5 points), highlighting that WebUI-to-Code performance on complex webpages still has considerable room for improvement.

\begin{figure*}[h]
  \includegraphics[width=\textwidth]{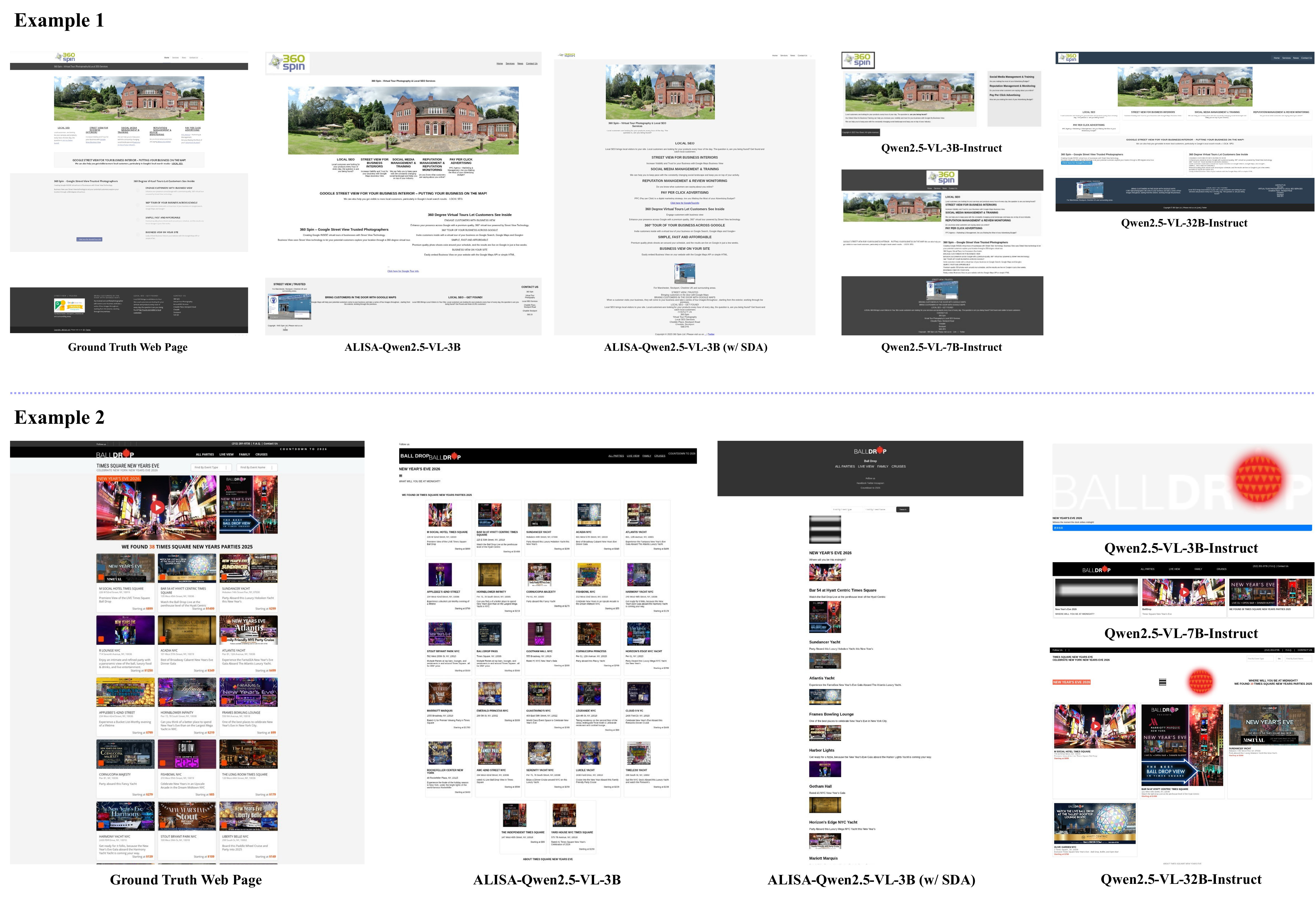}
  \caption{Comparison of HTML rendering results generated by different models from real webpage screenshots.}
  \label{a4}
\end{figure*}

\section{Case Study}

Figure~\ref{a4} presents two webpage examples comparing the outputs of models trained with the ALISA framework against Qwen2.5-VL-Instruct models of different sizes. Both ground-truth webpages contain a large number of elements, making them notably more complex than those in previous Web-related benchmarks. As a result, all models show certain discrepancies compared with the ground truth. Nevertheless, our ALISA-Qwen2.5-VL-3B achieves the highest layout consistency, accurately reproducing the horizontal alignment of the logo and navigation bar in both examples, as well as the grid layout of the main list in Example 2.

In contrast, the model trained solely with the SDA objective exhibits less consistent layouts, with visible mismatches compared with the default method. We also observe that vanilla Qwen2.5-VL models, regardless of parameter scale, often generate incomplete webpages. By comparison, both ALISA-Qwen2.5-VL-3B and ALISA-Qwen2.5-VL-3B (w/ SDA) produce more complete structures and significantly outperform Qwen2.5-VL-3B-Instruct. Moreover, larger models tend to produce more refined stylistic details, such as shadows and subtle visual effects, though some inconsistencies remain (for example, the navigation bar color in Example 1 generated by the 32B model differs from the target). These observations suggest that fine-tuning larger models could further enhance style alignment and overall rendering fidelity.

\end{document}